\documentclass[conference]{IEEEtran}
\IEEEoverridecommandlockouts
\usepackage{cite}
\usepackage{amsmath,amssymb,amsfonts}
\usepackage{algorithmic}
\usepackage{graphicx}
\usepackage{textcomp}
\usepackage{xcolor}

\usepackage[printonlyused,withpage,nolist,nohyperlinks]{acronym}
\usepackage{subfigure}
\usepackage{multirow}

\begin{acronym} \renewcommand{\\}{}  
\acro{RL}{Reinforcement Learning}
\end{acronym}

\def\BibTeX{{\rm B\kern-.05em{\sc i\kern-.025em b}\kern-.08em
    T\kern-.1667em\lower.7ex\hbox{E}\kern-.125emX}}
\begin{document}

\title{Replication of Impedance Identification Experiments on a Reinforcement-Learning-Controlled Digital Twin of Human Elbows
}

\author{\IEEEauthorblockN{1\textsuperscript{st} Hao Yu}
\IEEEauthorblockA{\textit{Edinburgh Centre for Robotics} \\
\textit{University of Edinburgh}\\
Edinburgh, The United Kingdom\\
s2273711@ed.ac.uk}
\and
\IEEEauthorblockN{2\textsuperscript{nd} Zebin Huang}
\IEEEauthorblockA{\textit{Edinburgh Centre for Robotics} \\
\textit{University of Edinburgh}\\
Edinburgh, The United Kingdom\\
s2057036@ed.ac.uk}
\and
\IEEEauthorblockN{3\textsuperscript{rd} Qingbo Liu}
\IEEEauthorblockA{\textit{Department of Mechanical Engineering} \\
\textit{University of Hong Kong}\\
Hong Kong, China\\
qingboliu0703@gmail.com}
\and
\IEEEauthorblockN{4\textsuperscript{th} Ignacio Carlucho}
\IEEEauthorblockA{\textit{School of Engineering and Physical Sciences} \\
\textit{Heriot-Watt University}\\
Edinburgh, The United Kingdom\\
ignacio.carlucho@hw.ac.uk}
\and
\IEEEauthorblockN{5\textsuperscript{th} Mustafa Suphi Erden}
\IEEEauthorblockA{\textit{School of Engineering and Physical Sciences} \\
\textit{Heriot-Watt University}\\
Edinburgh, The United Kingdom\\
m.s.erden@hw.ac.uk}
}


\maketitle

\begin{abstract}
This study presents a pioneering effort to replicate human neuromechanical experiments within a virtual environment utilising a digital human model. By employing MyoSuite, a state-of-the-art human motion simulation platform enhanced by \ac{RL}, multiple types of impedance identification experiments of human elbow were replicated on a musculoskeletal model. We compared the elbow movement controlled by an \ac{RL} agent with the motion of an actual human elbow in terms of the impedance identified in torque-perturbation experiments. The findings reveal that the \ac{RL} agent exhibits higher elbow impedance to stabilise the target elbow motion under perturbation than a human does, likely due to its shorter reaction time and superior sensory capabilities. This study serves as a preliminary exploration into the potential of virtual environment simulations for neuromechanical research, offering an initial yet promising alternative to conventional experimental approaches. An RL-controlled digital twin with complete musculoskeletal models of the human body is expected to be useful in designing experiments and validating rehabilitation theory before experiments on real human subjects.
\end{abstract}


\section{INTRODUCTION}
Over the past two decades, robotic technology has significantly advanced various aspects of neuromotor rehabilitation, including assessment, therapy, training, and motion assistance \cite{ReRobotBook2018}. These developments, aligning with digital healthcare trends, emphasise intelligent, digitised, and personalised approaches \cite{christou2022designing, gordon2023adaptive}. Although preliminary evidence suggested the superiority of robot-assisted rehabilitation, widespread clinical application is limited due to insufficient, convincing evidence from clinical trials \cite{RehabHandRobot2010, mehrholz2012electromechanical}. Therefore, it is essential to accumulate experimental data demonstrating the advantages of robot-assisted rehabilitation over traditional methods. This endeavour is challenging due to difficulties in patient recruitment, variability in disease manifestations, and the need for long-term monitoring \cite{ReRobotBook2018}.

With the advancement of simulation technologies, the digital twin of human musculoskeletal systems is introducing new solutions to address the challenge of early development and late-stage clinical validation of robot-assisted rehabilitation \cite{mazumder2023towards}. With a digital twin of the human body capable of acting like a human subject, it would be feasible to perform simulation experiments involving human and rehabilitation robots in virtual environments, reducing the dependence on real human experiments \cite{mcfarland2022musculoskeletal}. In the simulation software of human motion (\textit{e.g}., MyoSuite \cite{caggianoMyoSuiteContactrichSimulation2022a}, OpenSim \cite{seth2018opensim}, and AnyBody \cite{damsgaard2006analysis}), virtual models of human musculoskeletal systems are built with extensive experimental data of human anatomy and biomechanics. These musculoskeletal models simulate both mechanical and physiological properties of muscles, enabling detailed analyses of human motion, from muscular activities to kinetic features \cite{caggianoMyoSuiteContactrichSimulation2022a, mcfarland2022musculoskeletal}. Moreover, OpenSim can scale generic musculoskeletal models according to the dimensions of a subject to create a subject-specific digital twin \cite{seth2018opensim}, whereas MyoSuite supports physical interactions and computationally intensive simulations \cite{caggianoMyoSuiteContactrichSimulation2022a}. In the past, human digital twins realised in human motion simulators typically depend on real human motion data or muscle activity signals recorded with sensors in order to reconstruct motion in a virtual environment. This type of  human digital twin does not have motion control capabilities \cite{seth2018opensim}. In recent years, Reinforcement Learning (\ac{RL}) introduce a new solution for human motion virtualisation.

Compared to traditional simulation methods typically based on extensive data and prior knowledge, \ac{RL} is effective in complex, partially understood environments. Its strength lies in learning and adapting through environmental interaction, enabling nuanced and adaptive capture of human motor control \cite{schumacher2022dep, pengDeepMimicExampleGuidedDeep2018b, leeScalableMuscleactuatedHuman2019a}. Notable advancements have been made in simulating basic human movements, like point-reaching \cite{fischerReinforcementLearningControl2021b} and simple locomotion \cite{joosReinforcementLearningMusculoskeletal2020}, showcasing \ac{RL}'s capacity to emulate naturalistic human motion. This potential is further highlighted in the context of intricate tasks like hand-object manipulation \cite{bergSARGeneralizationPhysiological2023, caggianoMyoDexGeneralizablePrior2023b}. While \ac{RL} agents successfully emulate a range of human behaviours, what the \ac{RL} agents acquire from training is an optimal policy to achieve a motion task rather than human motor functions. The neuromotor characteristics of the policy need thorough validation and comparison with actual human motion. This aspect is crucial in bridging the gap between \ac{RL}-generated movements and their real-world human counterparts, ensuring a more accurate and realistic replication of human motor functions.

Impedance is a promising metric to evaluate the neuromotor similarities between RL agents and humans, because, in neuromechanics and rehabilitation robotics, it is widely used to describe the mechanical property of muscles and limbs and human somatosensory feedback in controlling posture and movement \cite{Burdet2013humanrobotics}. Numerous studies have been done on the impedance of human limb joints, so sufficient and reliable experimental data can be obtained from the prior studies as the ground truth of impedance identification in a virtual environment \cite{mizrahi2015impedance}. 

In this study, we realised the simulation of human-involved rehabilitation experiment in a virtual environment. Specifically, we reproduced the impedance identification experiments based on a virtual musculoskeletal model of a human elbow with an exoskeleton device in MyoSuite. The elbow model was controlled by an RL agent to complete holding and reaching tasks for identifying elbow impedance in static and dynamic conditions, respectively. The simulation results show that the \ac{RL}-controlled human musculoskeletal model partially mimics human's capability of stabilising posture and movement control. The successful reproduction of multiple classical impedance experiments in the virtual environment also illustrates the great potential of RL-based simulation of human musculoskeletal models in the field of rehabilitation robotics.

\section{METHODOLOGY}
\begin{figure}[tb]
\centering
     \includegraphics[width=0.45\textwidth]{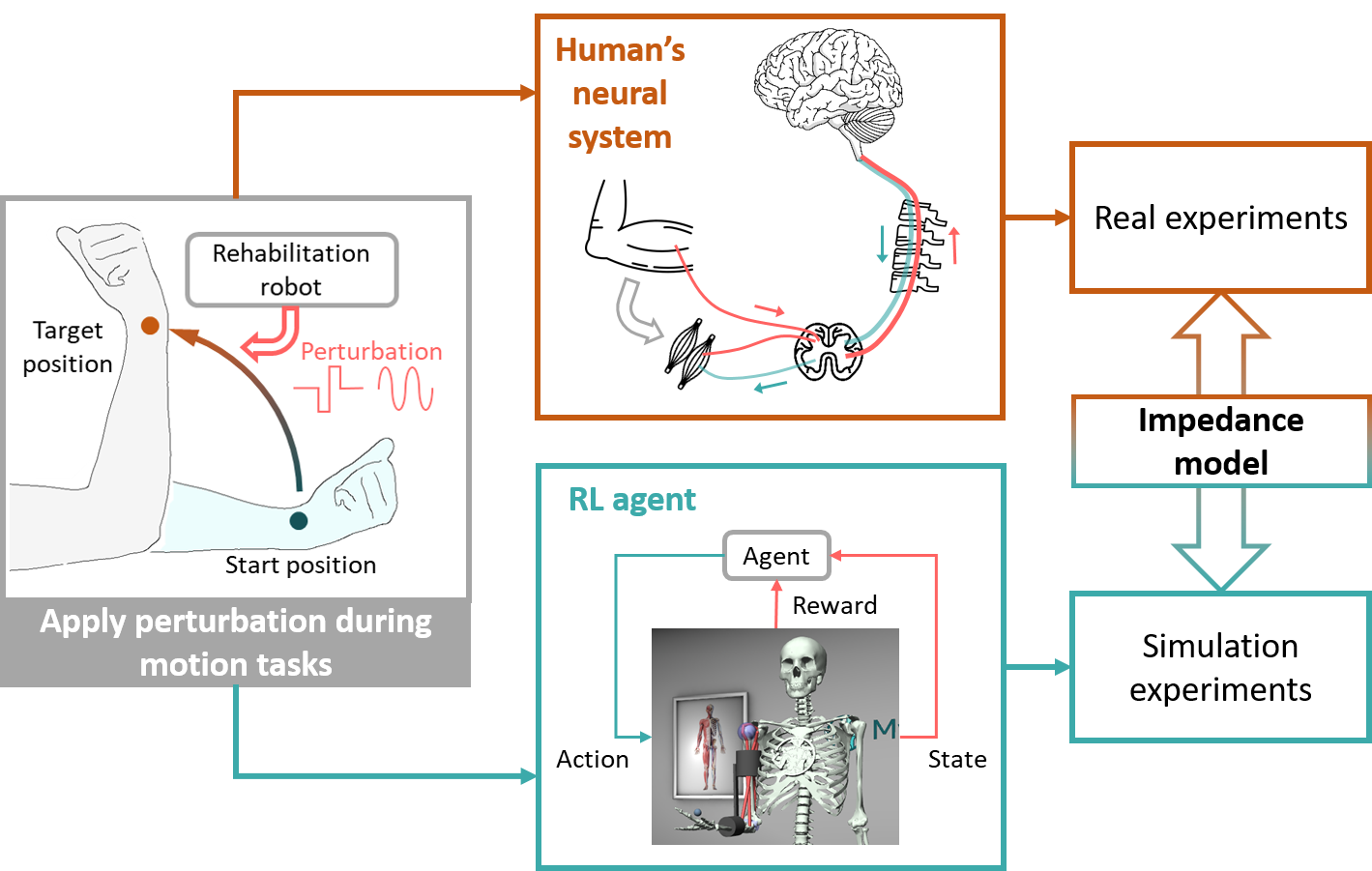}
\caption{Framework of comparing the neuromechanical properties between the RL-controlled elbow model and human subjects through the impedance identification experiments}
\label{fig:systemdiagram}
\vspace{-.4cm}
\end{figure}

In an impedance identification experiments, external torques are exerted by a rehabilitation robot on the elbow to observe human responses to perturbations during motion tasks, as illustrated in Fig. \ref{fig:systemdiagram}. In a real human experiment, the human responses to external perturbations are regulated by the motor neural system, whereas, in a simulation, the RL agent's actions are decided by its pre-trained policy. Central to the comparison between the human and the RL agent is the impedance model, which provides a quantitative analysis of the \ac{RL} agent's actions and human responses to perturbations. In this section, we first introduce the RL agent in MyoSuite. Then, we explain the elbow impedance identification method. 

\subsection{RL Agent}
In MyoSuite's framework, a \ac{RL} agent is trained to acquire a policy of motor control of a musculoskeletal model to execute human motion tasks within a MuJoCo-based environment \cite{todorov2012mujoco}. The policy controls the activations of the model's muscle actuators to interact with the environment, which alters the agent's state within the environment. The state encapsulates the model's current physical configuration, including muscle length and joint position. In response to each action, the environment provides a reward, a metric that evaluates the efficacy of the action in achieving motion tasks. The agent's policy, which directs its decision-making process of actions, is iteratively refined based on this reward feedback. This cycle of action, state observation, and reward evaluation, continuously updates the agent's policy to maximise overall reward, guiding the agent towards more effective strategies for the tasks in the simulated environment \cite{caggianoMyoSuiteContactrichSimulation2022a}.

The “myoElbowPose1D6MExoRandom-v0” environment of Myosuite is utilised in this work. This environment offers a biomechanical model tailored for the human right elbow. The elbow model is equipped with an exoskeleton model, which is able to apply perturbations on the elbow joint during movements. The actuators in the model consists of six elbow muscles (long head, lateral head, and medial head of triceps brachii, long head and short head of biceps brachii, and brachioradialis) and a motor of the exoskeleton. Correspondingly, the agent's actions are represented as a 7-dimensional vector, with each component ranging between -1 and 1. The first component of the action vector relates to the degree of freedom (DOF) in the exoskeleton, while the remaining six components are associated with the DOFs of the muscles. Moreover, the agent state is defined through 10 variables, encompassing the current simulation time, the joint angle of a single DOF in the exoskeleton, the joint velocity for that DOF measured in radians per second, activation values for each of the six muscle DOFs, and the difference vector between the current joint positions and target values for the exoskeleton's DOF.

The training algorithm employed is the natural policy gradient (NPG), a baseline sourced from the Reinforcement Learning Algorithms for MuJoCo Tasks (MJRL) \cite{Rajeswaran-NIPS-17, Rajeswaran-RSS-18}. The policy and specific training algorithm configurations are adopted from MyoSuite \cite{caggianoMyoSuiteContactrichSimulation2022a}. This established baseline has been replicated in our study to serve as a benchmark for subsequent experiments and comparisons. Detailed hyperparameters used in the training algorithm are presented in Table \ref{tab:npg_hyperparameters}.

\begin{table}[bt]
\centering
\caption{NPG Algorithm Hyperparameters}
\begin{tabular}{|c|c|}
\hline
\textbf{Hyperparameter}           & \textbf{Value}     \\ 
\hline
Policy Network      & MLP (32, 32)  \\
\hline
Value Network   & MLP (128, 128)\\
\hline
Batch Size & 64       \\
\hline
Epochs    & 2         \\
\hline
Learning Rate & 0.001 \\
\hline
Initial Log Std          & -0.25     \\
\hline
Minimum Log Std          & -1.0      \\
\hline
Discount Factor $\gamma$ & 0.995     \\
\hline
GAE $\lambda$            & 0.97      \\ 
\hline
Normalized Step Size $\delta$                & 0.1       \\
\hline

\end{tabular}
\label{tab:npg_hyperparameters}
\end{table}

\subsection{Elbow Impedance Identification}
\begin{figure*}[htb]
    \centering
         \includegraphics[width=0.9 \textwidth]{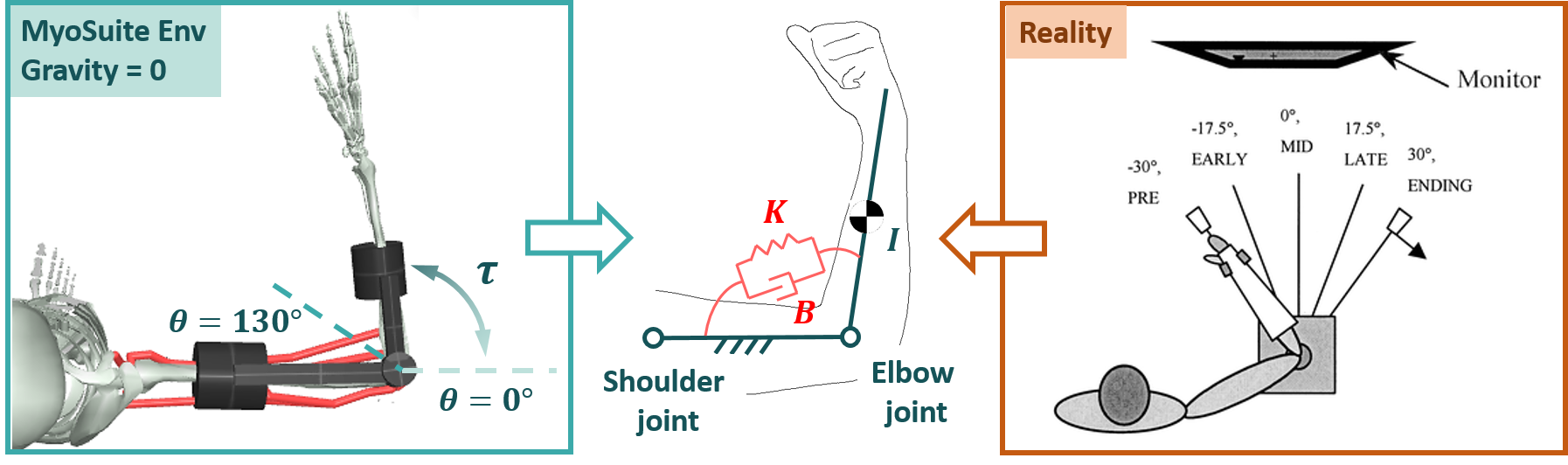}
    \caption{Neuromechanical characteristics of a virual or real elbow's responses to external perturbations can be described as an impedance model. Left: musculoskeletal model of an elbow in the MyoSuite from the top view. Middle: K-B-I impedance model of an elbow joint. Right: apparatus and protocol from Popescu et al.'s study on elbow impedance \cite{popescu2003elbow} ($\tau$ is the applied torque perturbation; $\theta$ is the elbow joint angle; $K$ is the joint stiffness; $B$ is the joint viscosity; $I$ is the moment of inertia of the forearm).}
    \label{fig:elbow}
    \vspace{-.4cm}
\end{figure*}

The dynamics of an elbow joint resisting to external perturbations can be mathematically described as a second-order linear system characterised by inertia, viscosity, and stiffness (see Fig. \ref{fig:elbow} and Eq. (\ref{Eq:impedance})) \cite{rack1973stretch, joyce1974forces, rack2011limitations}. These three parameters collectively represent the system's impedance to external perturbations. When a torque perturbation is applied to an elbow joint, its inertia creates a resistance perturbation proportional to acceleration. Simultaneously, the joint behaves like a spring and damper, generating two resistance proportional to the displacement and the velocity of movements, respectively. Consequently, the relationship between the external torque and these three types of resistance responses can be modelled as a linear K-B-I model: 
    \begin{align} \label{Eq:impedance} 
        \tau = I \ddot{\theta} + B \dot{\theta} + K \theta,
    \end{align}
where $\tau$ is the external torque perturbation; $K$ is the stiffness coefficient; $B$ is the viscosity coefficient; $I$ is the inertial coefficient; $\theta$ is the joint angle \cite{popescu2003elbow, selen2006impedance}. 

When the agonist and antagonist muscle pairs around the elbow joint maintain relaxed, the stiffness and viscosity are contributed primarily by the mechanical property of soft tissues \cite{Burdet2013humanrobotics}. In certain conditions, stretching the elbow joint triggers stretch reflexes in the motor neurons, and potentially long-latency reflexes involving the central nervous system. These reflexes serve as a self-protection mechanism for the muscles and enhance the stability of bodily movements. They introduce reflex-related muscle forces into the resistance against external perturbations \cite{rack2011limitations}. 

For the linear K-B-I model, various system identification methods designed for linear time-invariant systems can be employed to estimate the model parameters. This process involves stimulating the system with specific inputs and recording the outputs to estimate the model parameters, based on the response characteristics of the system in both the time and frequency domains \cite{shinners1998controltheory}. To identify the K-B-I model of a human elbow, prior studies have utilised the sinusoidal displacement perturbations \cite{rack1973stretch, joyce1974forces}, pseudo-random torque/displacement disturbances \cite{selen2007fatigue, bennett1992time}, and bi-polar and symmetric torque pulses \cite{popescu2003elbow, selen2006impedance} as input signals.  For torque-based input scenarios, the system output is typically the change in joint angle \cite{popescu2003elbow, selen2006impedance}.

The modelling and identification approach for the musculoskeletal system of an elbow, as described above, has been reliably validated in many studies, establishing it as a reliable method for analysing the neuromotor control of the elbow \cite{rack1973stretch, joyce1974forces, popescu2003elbow, selen2006impedance}. In this study, the same approach is applied to assess the motor function of a virtual elbow controlled by an RL agent. The impedance parameters of real human elbows from the prior studies serve as the ground truth for our analysis. This data is compared with the impedance parameters of the digital twin of human elbow identified in the same experimental conditions. 

\section{Experiment Protocol}
The simulation experiment was designed following the experimental protocol of prior research on elbow impedance~\cite{popescu2003elbow, selen2006impedance}. In these studies, subjects were seated with their upper torso restrained. Both the upper arm and forearm of the subject were fixed onto two segments of a single-joint linkage on a plane. This linkage's pivot aligns parallel to the elbow joint, typically driven by a motor, allowing for both active and passive flexion or extension of the subject's elbow without gravitational influence. In our work, we replicated this experimental protocol in MyoSuite using an elbow model with a rigid exoskeleton, as shown in Fig.~\ref{fig:elbow} and then conducted four distinct types of impedance identification experiments. 

\subsection{Inertia Estimation}
We first validated the perturbation-based impedance identification in the MyoSuite environment through inertia estimation (IE) experiments. The total inertia of the subject's forearm and hand was estimated using a 20-second transient sinusoidal torque input. This input featured an increasing frequency range from 2 to 20 $Hz$ and a constant magnitude of 10 $N{\cdot}m$. One trial was done for each frequency setting. During these trials, the start position of the elbow joint was set to 0.773 $rad$, coinciding with the equilibrium point of the agonist and antagonist muscles when inactive. All muscle actuators and the exoskeleton motor were deactivated for these experiments. The inertia was then calculated from the joint displacement response, employing the algorithm described in \cite{popescu2003elbow}. This method is designed to isolate the inertia component by eliminating the contributions of reflexes and viscosity to the displacement response  \cite{rack2011limitations}. The inertia values derived from our experiments were then validated with the internal parameters predefined in the model file.

\subsection{Mechanical Impedance Identification}
In addition to employing sinusoidal torque, the identification method based on bi-polar and symmetric torque pulses was validated in mechanical impedance identification (MII) experiment. There were ten trials conducted in this experiment. In each trial, we exerted a brief, bi-polar, and symmetric torque pulse to identify the viscosity and stiffness of the elbow joint at its equilibrium point. The torque perturbation is a bi-directional pulse, lasting 0.05 $s$ at each direction (0.1 $s$ in total), with an amplitude of 20 $N{\cdot}m$. The estimation algorithm used in this experiment adhered to the method presented in \cite{popescu2003elbow, selen2006impedance}. This experiment also helped reveal the the pure mechanical property of the elbow model, offering crucial comparative data for subsequent experiments involving reflex actions and dynamic movements.

\subsection{Static Impedance Identification}
After validating the identification method, we then activated the muscle actuators, which were under the control of an RL agent, to enable position control of the virtual elbow. Previous research has indicated that elbow impedance varies across different movement states \cite{popescu2003elbow, selen2006impedance, selen2007fatigue, bennett1992time}. In light of this, we evaluated the responses of the virtual elbow to external perturbations in both static and dynamic states to understand these variations comprehensively.

In the static impedance identification (SII) experiment, the RL agent modulated the activation levels of each muscle to maintain the elbow joint at a specific target position, countering the effects of bi-polar and symmetric torque pulses (0.05 $s$ for each direction with an amplitude of 20 $N{\cdot}m$). The target position was set to range from 0 to 150$^\circ$, incrementing in steps of 10$^\circ$. For each target position, five trials were conducted.

\subsection{Dynamic Impedance Identification}
The last experiment, dynamic impedance identification (DII) experiment, focused on how the RL agent modulates elbow impedance during a dynamic movement task. Initially, in the first 10 trials, the RL agent flexed the elbow from 0$^\circ$ to 150$^\circ$ and then extended it back to the start position. No external torque was applied in these trials to capture the reference movement profiles. Subsequently, the same flexion and extension movements were repeated for additional 10 trials, but this time under the influence of a torque pulse (0.05 $s$ for each direction and 40 $N{\cdot}m$). The difference between the reference and interfered movements were calculated and used as the output for impedance identification, providing a clear measure of the RL agent's impact on dynamic movement modulation.

\section{Data Analysis}
The inertia was estimated based on the frequency-amplitude characteristics of the sinusoidal torque inputs and displacement outputs. In the IE experiment, limb inertia emerged as the dominant response component under high-frequency sinusoidal perturbation \cite{rack2011limitations}. Given that, the impedance model, as detailed in Eq. \ref{Eq:impedance}, was streamlined into a purely inertial model, represented by the following equations:
    \begin{align} 
        \tau_s (t) = I \ddot{\theta_s} (t), \\
        \tau_s (t) = T_s \sin{(2 \pi f t)}, \\
        \theta_s (t) = A_s \sin{(2 \pi f t)},
    \end{align}
where $\tau_s$ is the applied sinusoidal torque; $\theta_s$ is the displacement of the elbow joint relative to the start position; $T_s$ and $A_s$ are the amplitudes of the input and output; $f$ is the frequency of the torque and displacement; $t$ is the time variable. The second derivative of $\theta_s$ is
    \begin{align} \label{Eq: angle_derivatives} 
        \ddot{\theta_s} (t) = - A_s (2 \pi f )^2 \sin{(2 \pi f t)}.
    \end{align}        
With the Eq. (2)-(5), the inertia can be calculated by
    \begin{align} \label{Eq: inertia_estimation} 
        I = T_s/X, \\
        X = 4 \pi^2 f^2 A_s.
    \end{align}      
The main frequency and amplitude of both the inputs and outputs were extracted using a Finite Fourier Transform. This allowed for the estimation of inertia by analysing the slope of the curve relating the torque amplitude to the product of the position amplitude and the squared frequency. After validated with the internal model parameter, the calculated inertia was fixed and utilised in the data analysis of the following experiment.

To estimate the stiffness and viscosity, the impedance parameters were adjusted to minimise the error between the simulated and predicted traces based on the ordinary differential equation (ODE) of the K-B-I model: 
    \begin{align} \label{Eq:ODE_KBI} 
        \tau(t) = I \frac{d^2 \theta}{d^2 t} + B \frac{d \theta}{d t} + K \theta.
    \end{align}  
Given the initial displacement, denoted as $\theta_0 = 0$, and the profile of the exerted torque, $\tau(t)$, the displacement response was predicted by numerically integrating the ODE over the duration of the perturbation. The optimal values of stiffness and viscosity, K and B, were searched by globally optimising the error function between the simulated and predicted traces:
    \begin{align} \label{Eq: global_opt} 
        \theta_p (\tau(t)) = \theta_p (K, B, I, \tau(t)), \\
        \triangle_{min} = \underset{K,B}{min} |\theta_p (\tau(t))-\theta_r(t)|,
    \end{align} 
where $\theta_p$ and $\theta_r$ are the trace predicted by the model and the real trace registered in a simulation trial, respectively. The global optimisation algorithm was realised with the differential evolution function provided by the SciPy library. This optimisation process ensured the closest match possible between the model's behaviour and the RL agent's responses.

For the MII and SII experiments, the real trace, $\theta_r(t)$, was the mean of the traces recorded in multiple trials. For the DII experiment, the mean traces of the reference movements and the interfered movements were respectively calculated. The difference between the two mean traces was used as the real trace in Eq. (\ref{Eq: global_opt}).

\section{RESULTS}
In this section, we present key findings from our experiments on neuromechanical behaviours using the MyoSuite model. We accurately estimated the moment of inertia around the elbow joint in the IE experiments, and also found that the K-B-I model's predictions closely mirrored actual simulation results in the MII experiments. These two experiment results show that the perturbation-based impedance identification methods are appliable for the MyoSuite model. Following that, in the SII and DII experiments, the RL agent effectively stabilised the elbow movement against torque perturbations, demonstrating greater impedance than a human. 

\subsection{Moment of Inertia}
\begin{figure}[tb]
    \centering
    \includegraphics[width=0.5\textwidth]{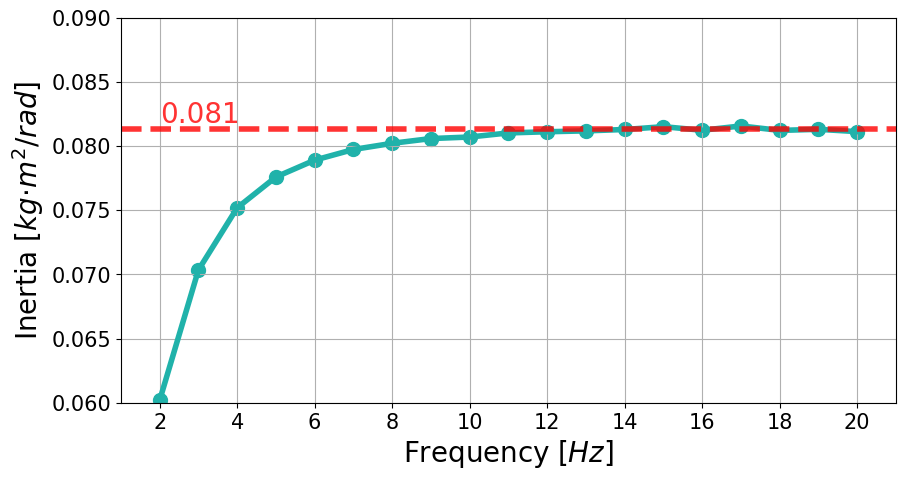}
    \caption{Estimation of the moment of inertia based on the MyoSuite elbow model; the green line with dots is the estimate results in the simulation experiments; the red line is the ground truth calculated according to the model parameter}
    \label{fig:inertia}
    \vspace{-.2cm}
\end{figure}

Accurate estimation of inertia is crucial in the identification of impedance, particularly when the applied perturbation is fast or of high frequency. This is because any error in estimating inertia could significantly affect the estimations of stiffness and viscosity \cite{popescu2003elbow}. In the model library of MyoSuite, the inertia can be extracted from the model's source file. Specifically, for the elbow model used in our study, the total moment of inertia around the elbow joint had to be calculated based on the mass and inertia matrix of the forearm, which amounts to 0.081 $kg{\cdot}m^2/rad$. This internal parameter of the model serves dual purpose: firstly, as a baseline (the red dashed line in Fig. \ref{fig:inertia}) for the IE experiment; and secondly, as a known parameter in subsequent experiments aimed at identifying $K$ and $B$.

In Fig. \ref{fig:inertia}, it is observed that the inertia estimation converges towards the baseline as the frequency increases. Notably, the estimated value of inertia stabilises around the ground truth line beyond the 10 $Hz$ mark. This indicates a consistent match between the simulation results and the actual parameters of the model. 

\subsection{Elbow Impedance in the Static Condition}
\begin{figure}[tb]
    \centering
         \includegraphics[width=0.5\textwidth]{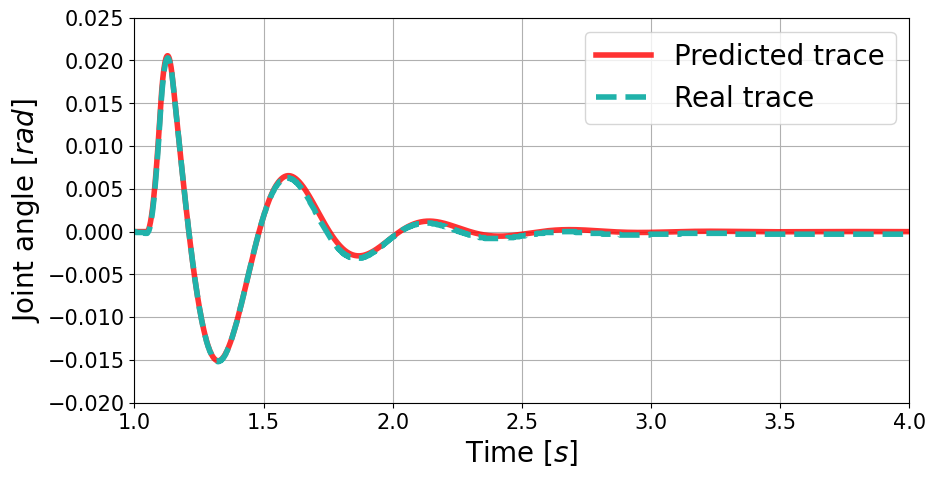}
    \caption{Resulting joint angle trace in an example of the MII experiment and corresponding predicted trace generated by the identified K-B-I model.}
    \label{fig:real_predicted}
\end{figure}

\begin{figure}[tb]
    \centering
         \includegraphics[width=0.5\textwidth]{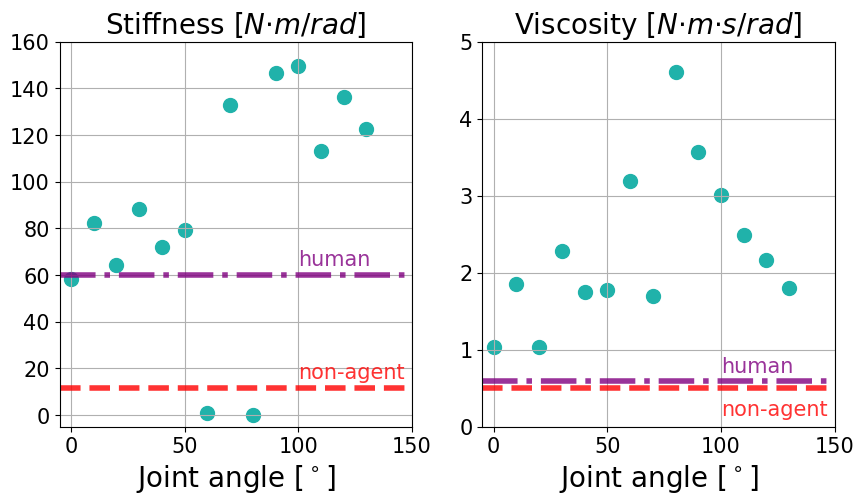}
    \caption{Elbow stiffness and viscosity of the human and RL agent in the static condition; the green points are the identification results of the SII experiment; the red dashed lines of non-agent are the identification results of the MII experiment; the slash-and-dash purple lines are the experimental results from human subjects \cite{popescu2003elbow}.}
    \label{fig:static_impedance}
    \vspace{-.4cm}
\end{figure}

The identification of stiffness and viscosity based on the pulse torque perturbation has been validated in multiple prior studies, but it is being applied for the first time in a virtual environment and digital human model \cite{popescu2003elbow, selen2006impedance}. Therefore, it is essential to verify whether the K-B-I model can reliably represents the behaviour of a virtual human joint and if the perturbation method is appropriately configured and effective within the MyoSuite environment. An illustrative example of this verification process is provided in Fig. \ref{fig:real_predicted}, which displays a comparison between the actual position trace recorded in the simulation and the predicted trace generated by the K-B-I model. The model's predicted traces closely mirror the actual traces from the simulation experiments when a disturbance is introduced. The root mean squared error between the predicted traces of the model and the actual traces of the simulation experiments is 11.65e-4 $rad$. The good prediction results illustrate the effectiveness of the identification method. The identification results of the MII experiment were $K=11.65\ N{\cdot}m/rad$ and $B=0.50\ N{\cdot}m/rad$ (the red dashed lines in Fig. \ref{fig:static_impedance}). In the MII experiment, the muscle actuators of the model were disabled, so no agent contribute to the impedance.

This impedance identification method verified as above was used across the entire range of elbow motion in the SII experiment, leading to the results depicted in Fig. \ref{fig:static_impedance}. In this figure, the baselines for human elbow impedance are from the mean values reported in \cite{popescu2003elbow}, which are $K=60\ N{\cdot}m/rad$ and $B=0.60\ N{\cdot}m/rad$ (the purple slash-and-dashed lines in Fig. \ref{fig:static_impedance}).  It is evident from these results that the stiffness of the elbow in response to external impulsive torque disturbances significantly increases when either neural reflexes or an RL agent are incorporated into the motor control mechanism. The stiffness of the elbow under RL agent control is higher than that of human subjects. The stiffness value increases from about 60 $N{\cdot}m/rad$ to 140 $N{\cdot}m/rad$ with the rise of the joint angle. Additionally, the joint viscosity of the elbow controlled by the RL agent is also observed to be higher than that of a real human elbow. In contrast, the mechanical viscosity of the uncontrolled model joint aligns closely with the baseline viscosity derived from real human data.

\subsection{Elbow Impedance in the Dynamic Condition}
\begin{figure}[tb]
    \centering
         \includegraphics[width=0.5\textwidth]{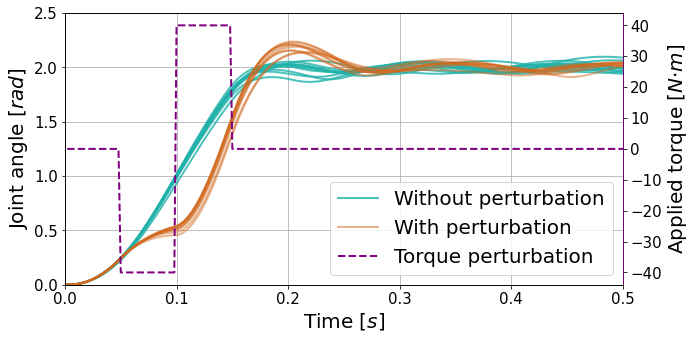}
    \caption{Traces of the elbow movements with or without the torque perturbation during flexion; the solid lines are the movement traces in the simulation; the dashed line is the torque profile applied during the movements}
    \label{fig:dynamic_movement}
\end{figure}

Apart from maintaining a static position, the stability of movement in dynamic environments is a crucial characteristic for evaluating human motor control \cite{selen2006impedance}. In the DII experiment, the RL agent effectively stabilised the elbow's movement to reach the set target position, even under the influence of torque perturbation, as illustrated in Fig. \ref{fig:dynamic_movement}. The motion of the elbow joint was initially impeded by the disturbance, causing a deviation from the reference trace observed in the absence of disturbance. The RL agent then responded adaptively to this disturbance through feedback control mechanisms. Compared to the reference movement, the disturbed trace slightly surpassed the target position by 0.2 $rad$ but subsequently corrected itself, settling at the intended target. This process took a marginally longer time, approximately 0.05 $s$ more than the reference trace. The impedance parameters of the elbow during this dynamic movement were calculated and are detailed in Table II.

\begin{table}[tb]
    \vspace{-.2cm}
    \centering
    \caption{Elbow impedance in the dynamic conditions.}
    \begin{tabular}{|c|c|c|c|c|}
    \hline
    \multirow{2}{*}{\textbf{Studies}} & \multirow{2}{*}{\textbf{Movement}} & \textbf{Stiffness} & \textbf{Viscosity}\\
    ~ & ~ & {\scriptsize$[N{\cdot}m/rad]$} &  {\scriptsize[$\times10^{-2}N{\cdot}m{\cdot}s/rad$]} \\
    \hline
    \multirow{2}{*}{Ours} & Extension & 28.41 & 2.41 \\
                    \cline{2-4} 
                    & Flexion & 50.83 & 2.71 \\
    \cline{1-4} 
    \cite{popescu2003elbow} & Reaching Task & 0 & $0.5\sim1$ \\
    \hline
    \cite{bennett1993torques} & Slow Rotation & 10 & $-$ \\
    \hline
    \cite{xu1999robust} & Cyclical & 20 & $1\sim2$ \\
    \hline
    \end{tabular}
    \label{impedance_table}
    \vspace{-.4cm}
\end{table}

\section{Discussion}
Leveraging the human motion physics simulation environment offered by MyoSuite, along with the detailed human physiological and anatomical characteristics embedded in the model, we successfully replicated joint impedance identification experiments in a simulation setting. By estimating the inertia of the upper arm and the purely mechanical impedance of the elbow joint, we find that the impedance identification method, previously validated in real human subjects, is equally applicable and valid for musculoskeletal models within a simulation environment (Fig. \ref{fig:inertia} and \ref{fig:real_predicted}). Moreover, given the higher impedance values, the performance of the RL-based agent in controlling the elbow joint under external disturbances is more stable than human neuromotor control (Table I).

\subsection{Identification of Impedance in a Virtual Environment}
It is important to note that the linear K-B-I model simplifies the non-linear properties of the musculoskeletal system in the elbow. Therefore, the parameters identified using the methods designed for linear time-invariant systems need careful interpretation, given the potential discrepancies with the more complex real-world behaviour of human joints \cite{rack1973stretch, rack2011limitations}. The implications of experimental conditions and the specific settings of perturbations play a critical role and must be thoroughly considered when discussing the impedance parameters.

The use of sinusoidal inputs with sufficient bandwidth and suitable amplitude for system identification is an effective approach for a second-order linear system, including the simplified elbow joint system, because it gives a more complete picture of the system's characteristics in both the time and frequency domains compared to pulse inputs \cite{shinners1998controltheory}. We used a sinusoidal torque input to successfully identify the moment of inertia of the model's forearm. The experiment resulted in an inertia measurement aligning with the anthropometric data of an adult male as well \cite{zatsiorsky1985estimation}. However, the sinusoidal-torque perturbation was not utilized in subsequent impedance identification experiments. This decision was based on two key factors: firstly, reliable experimental data for elbow impedance identification available in the literature primarily pertains to sinusoidal-displacement perturbation methods \cite{rack1973stretch, joyce1974forces}. Secondly, the MyoSuite model used in this study is not configured to accommodate fast sinusoidal displacements, thereby limiting the feasibility of using this specific type of perturbation for our experiments.

In addition to sinusoidal disturbances, torque pulses are also viable for evaluating joint impedance in a virtual environment, though the selection of pulse duration requires careful consideration. Due to the refresh rate of the MyoSuite environment, any applied pulse perturbation must have a duration exceeding 0.02 $s$. Our experimentation reveals that pulse perturbations function reliably when their duration is approximately twice that of the refresh rate, i.e., 0.04 $s$. This constraint impedes the ability to precisely replicate certain experimental conditions from previous human studies within the simulation environment \cite{popescu2003elbow}.

\subsection{Comparison between Human and RL-based control}
The elbow joint controlled by the RL agent exhibited higher impedance to external disturbances than human experimental data in both static and dynamic motion tasks. In the static motion experiment, the impedance data of the RL agent's elbow were slightly higher than those of humans, but remained within the same order of magnitude. In contrast, during dynamic motion experiments, the human elbow impedance tended to approach to zero, while the stiffness of the RL agent's elbow was approximately halved, and its viscosity remained nearly consistent with that observed in the static motion experiment.

This difference is likely linked to the reaction time. The controller refresh rate (the cycle time for a single control loop) for the musculoskeletal model managed by the RL agent is 0.02$s$, faster than human short-latency and long-latency reflexes \cite{erden2015hand}. Therefore, in response to fast and short pulse disturbances, the RL agent can more swiftly detect displacement caused by the disturbance and adjust joint impedance to stabilise the movement. Conversely, a human subject would require more time to react and counteract the disturbance. This is supported by past studies where the human elbow joint has demonstrated greater impedance in slower or cyclical movements. Slower movements provide longer reaction time, while cyclical movements are predictable, so subjects can show better stabilisation capabilities in these experiments  \cite{bennett1993torques, xu1999robust, popescu2003elbow}.

In addition, the somatosensory capabilities of the RL agent in the MyoSuite simulation are superior to those of humans. The RL agent can instantly know the exact angle of the joint after each control cycle refresh, whereas the human body relies on processing information from intramuscular and tactile receptors, as well as visual information, to understand elbow position \cite{Burdet2013humanrobotics}. Unless the body's stretch reflexes are triggered or the perturbation is anticipated, humans are generally incapable of executing short-latency reactions to external disturbances. The RL agent, equipped with superior sensory abilities, can rapidly discern and respond to changes in joint position after each control cycle. This heightened sensory acuity enables the RL agent to achieve greater stabilisation, which is reflected in the higher impedance observed in the simulation experiments.

\section{Conclusion}
This paper represents the first effort to replicate classical human biomechanical experiments on a digital person within a virtual environment. This endeavour explores the feasibility of validating and conducting biomechanical simulation experiments in virtual settings in a manner that is both rapid and cost-effective. It also sheds light on the similarities and differences between the simulation outcomes derived from the current state-of-the-art human motion simulation environment, MyoSuite, and the control mechanisms of actual human motion. The comparison between the two is based solely on impedance as a metric. Future work will expand this comparison to include electromyography signals as another crucial evaluative metric, offering a more comprehensive understanding of neuromuscular control and interaction within simulated environments.

\section*{ACKNOWLEDGMENT}
This study was supported by EPSRC Centre for Doctoral Training in Robotics and Autonomous Systems under the Grant Reference EP/S023208/1. 


\bibliographystyle{IEEEtran}
\bibliography{ref2}

\addtolength{\textheight}{-12cm}   


\end{document}